\pdfoutput=1

\documentclass[11pt]{article}

\usepackage[final]{acl}

\usepackage{times}
\usepackage{latexsym}

\usepackage[T1]{fontenc}

\usepackage[utf8]{inputenc}

\usepackage{microtype}

\usepackage{inconsolata}

\usepackage{graphicx}

\usepackage{algorithm}
\usepackage{algorithmic}
\usepackage{amsmath}

\renewcommand{\algorithmiccomment}[1]{\bgroup\hfill//~#1\egroup}

\usepackage{booktabs} 
\usepackage{multirow} 
\usepackage{pifont}

\usepackage{xcolor}

\title{FLRC: Fine-grained Low-Rank Compressor for Efficient LLM Inference}

\author{
 \textbf{Yu-Chen Lu\textsuperscript{1,2}},
 \textbf{Chong-Yan Chen\textsuperscript{1}},
 \\
 \textbf{Chi-Chih Chang\textsuperscript{3}},
 \textbf{Yu-Fang Hu\textsuperscript{1}},
 \textbf{Kai-Chiang Wu\textsuperscript{1}}
\\
 \textsuperscript{1}National Yang Ming Chiao Tung University,
 \\
 \textsuperscript{2}Macronix International Co., Ltd.,
 \textsuperscript{3}Cornell University
\\
 \small{
   \textbf{Correspondence:} \href{mailto:yuchen.cs11@nycu.edu.tw}{yuchen.cs11@nycu.edu.tw}
 } 
}

\begin{document}
\maketitle

\begin{abstract}

Although large language models (LLM) have achieved remarkable performance, their enormous parameter counts hinder deployment on resource-constrained hardware. Low-rank compression can reduce both memory usage and computational demand, but applying a uniform compression ratio across all layers often leads to significant performance degradation, and previous methods perform poorly during decoding. To address these issues, we propose the \emph{Fine-grained Low-Rank Compressor (FLRC)}, which efficiently determines an optimal rank allocation for each layer, and incorporates progressive low-rank decoding to maintain text generation quality. Comprehensive experiments on diverse benchmarks demonstrate the superiority of \textit{FLRC}, achieving up to a 17\% improvement in ROUGE-L on summarization tasks compared to state-of-the-art low-rank compression methods, establishing a more robust and efficient framework to improve LLM inference.

\end{abstract}
\begin{figure}[t]
  \includegraphics[width=\columnwidth]{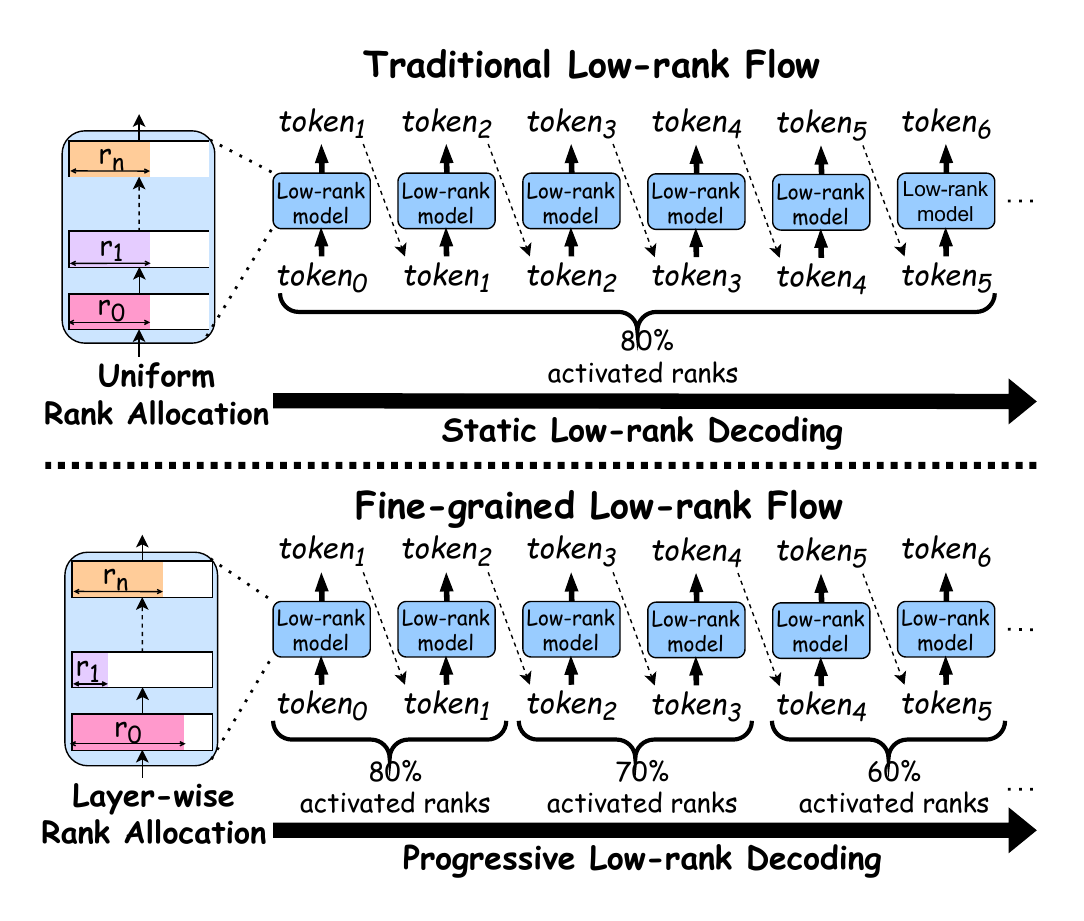}
  \caption{The differences between FLRC and traditional low-rank compression. As shown on the left side of the figure, we can determine the optimal number of ranks to preserve for each layer. On the right side, during the decoding stage, our approach gradually reduces the model's overall activated rank as more tokens are generated, unlike previous static methods, thereby decreasing the parameter usage and computational requirements while maintaining the quality of the generated output.}
  \label{fig:progressive}
\end{figure}
\section{Introduction}
In recent years, large language models (LLM) \citep{zhang2022opt, touvron2023llama, jiang2023mistral, liu2024deepseek} have achieved remarkable progress in text understanding and generation, finding widespread applications in areas ranging from customer service to data analysis. However, the substantial parameter counts and high computational demands of these models pose significant challenges for deployment in resource-constrained environments such as mobile devices and edge servers.

To address these challenges, various model compression techniques have been proposed to reduce the computational and memory requirements of LLM while maintaining performance. Notable methods include model pruning \citep{ma2023llm, akhauri2024shadowllm} and quantization \citep{shao2023omniquant, liu2024spinquant}. Among these, low-rank compression methods based on singular value decomposition (SVD) \citep{yuan2023asvd, wang2024svd} have shown particular promise in reducing both model size and computational cost.

Despite their potential, low-rank compression methods face several challenges that must be addressed. First, each layer (and even each projection) has its own tolerance for compression (c.f. Appendix~\ref{sec:importance}). Previous studies \citep{lin2024modegpt, ji2024feature, shao2024one} have attempted to assign different, optimal compression rates to each component, but these methods are often time-consuming or insufficiently precise. Another significant issue is that prior work primarily evaluates compressed models on prefill-centric benchmarks, such as perplexity or common-sense reasoning tasks, which are limited to single-token generation. Our analysis reveals that even state-of-the-art SVD-based methods suffer from notable accuracy degradation on tasks that require multiple decoding iterations, such as text summarization.

 As shown in Figure~\ref{fig:progressive}, we propose \emph{Fine-grained Low-Rank Compressor (FLRC)} to overcome current limitations. Our framework introduces two key innovations. First, we develop an efficient, gradient-based rank allocation algorithm that is significantly faster and more accurate than existing methods. Second, we implement a dynamic low-rank compression paradigm that adjusts the rank allocation during each token generation, starting with a conservative compression rate and progressively increasing it to maintain high accuracy at the same overall compression ratio.

Experimental results on popular LLaMA model families further validate our approach. In our experiments, our rank allocation algorithm reduces search time by up to 49× compared to previous methods, and \emph{FLRC} achieves up to a 17.35\% higher ROUGE-L score on summarization benchmarks, setting a new standard for efficient and accurate model compression.

\section{Related Works}

Low-rank compression \citep{kaushal2023lord, hsu2022language} has emerged as an effective strategy for reducing both parameter counts and computational overhead in neural networks. ASVD \citep{yuan2023asvd} mitigates the impact of outlier activations by scaling weight matrices based on activation distribution. Additionally, it introduces a rank allocation strategy to assign appropriate parameters ratio to each layer. However, this search method is extremely time-consuming. In contrast, our proposed rank search significantly reduces search time and, under high compression rates, finds rank allocation that deliver superior performance.

Another related work, SVD-LLM \citep{wang2024svd}, introduces a truncation-aware data whitening method to better correlate singular values with compression errors, allowing the truncation of smaller singular values with minimal impact on error. However, despite these improvements, many low-rank compression methods still perform suboptimally during the decoding phase of LLM inference. To overcome this limitation, we propose progressive low-rank decoding, which maintains high text generation quality even under aggressive compression, thereby improving the practicality of compressed LLM in real-world generation tasks.
\begin{algorithm}[!bht]
\caption{Layer-wise Rank Allocation}
\label{alg:rank_search}
\begin{algorithmic}[1]

\REQUIRE{\begin{minipage}[t]{\linewidth}
    Model $M$ with layers $L$, \\
    where each layer $l \in \mathcal{L}$ \\
    contains a set of projections $P_l$; \\
    Calibration dataset $\mathcal{D}$; \\
    Rank budget target $R_{\text{budget}}$.
    \end{minipage}}

\ENSURE {
    Rank allocation $\{r_{l,p}\}_{l \in \mathcal{L},p \in P_l}$.
    }

\STATE 
    $\{\,\mathbf{G}_{l,p}\}\; \gets \text{ComputeGradient}(M, \mathcal{D})$ 
    
\FOR{each layer $l \in \mathcal{L}$}
   \FOR{each projection $p \in P_l$}
        \STATE
        Compute the importance: \\
        $
            \alpha_{l,p} \;=\; \sum_{i}\Bigl(\mathbf{G}_{l,p}[i] \times \mathbf{W}_{l,p}[i]\Bigr)^2.
        $
    \ENDFOR
\ENDFOR

\STATE Compute the total importance: \\
    $
      S \;=\; \sum_{l \in \mathcal{L}} \sum_{p \in P_l} \alpha_{l,p}.
    $

\FOR{each layer $l \in \mathcal{L}$}
   \FOR{each projection $p \in P_l$}
        \STATE Allocate rank proportionally: \\
        $
          r_{l,p} \;=\; \mathrm{round}\Bigl(\frac{\alpha_{l,p}}{S} \times R_{\text{budget}}\Bigr).
        $
    \ENDFOR
\ENDFOR

\STATE \textbf{return} $\{r_{l,p} \mid l \in \mathcal{L}, p \in P_l\}$.

\end{algorithmic}
\end{algorithm}
\section{Proposed Method}

Our proposed \emph{Fine-grained Low-Rank Compressor (FLRC)} consists of two main components.

\subsection{Fisher-based Layer-wise Rank Allocation}
In LLM, different weight matrices---and even different projections within the same layer---exhibit varying capacities to tolerate compression. A uniform compression ratio across all layers can thus be suboptimal, as it may overcompress some components while undercompressing others. To address this issue, we propose the \emph{Fisher-based Layer-wise Rank Allocation} algorithm, which computes an optimal rank allocation for each projection, preserving crucial projection ranks while effectively reducing overall model size. An overview of our algorithm is provided in Algorithm~\ref{alg:rank_search}.

Our method begins by passing a calibration dataset $\mathcal{D}$ through the model $M$ and computing the gradients via backward propagation. Let $L$ denote the set of all layers in the model, and for each layer $l \in L$, let $P_l$ be the set of projections in that layer. For each layer $l \in L$ and each projection $p \in P_l$, we denote the corresponding weight vector as $\mathbf{W}_{l,p}$ and its gradient as $\mathbf{G}_{l,p}$. We then calculate a fisher-based \citep{abdelfattah2021zero} importance value $\alpha_{l,p}$, defined as:
\begin{equation}
\label{eq:alpha}
\alpha_{l,p} \;=\; \sum_{i}\Bigl(\mathbf{G}_{l,p}[i] \times \mathbf{W}_{l,p}[i]\Bigr)^2,
\end{equation}
which measures the sensitivity of each projection by incorporating both the gradient and the weight values. Higher $\alpha_{l,p}$ values indicate that the projection is more critical and should be compressed less aggressively (or potentially left uncompressed), whereas lower values suggest that the projection can tolerate more aggressive compression. For further details on Equation~\ref{eq:alpha}, please refer to Appendix~\ref{sec:metrics}.

After computing the importance values for all projections, we sum them to obtain the total importance score $S$.
We then allocate the rank for each projection proportionally to its importance by setting:
\begin{equation}
\label{eq:rank_allocation}
r_{l,p} \;=\; \mathrm{round}\Bigl(\frac{\alpha_{l,p}}{S} \times R_{\text{budget}}\Bigr),
\end{equation}
where $R_{\text{budget}}$ is the overall rank budget target, adjustable based on the desired level of overall parameter compression. This yields a layer-wise rank allocation $\{r_{l,p} \mid l \in \mathcal{L}, p \in P_l\}$ that specifies the number of ranks retained for each projection in each layer, reflecting their relative importance.

This adaptive strategy ensures that the available compression budget is efficiently distributed across the model, focusing more resources on the most impactful components. As a result, our rank allocation method achieves a better balance between compression and performance compared to methods that apply a uniform compression ratio across all layers.

\subsection{Progressive Low-rank Decoding}

In text generation tasks, earlier tokens play a more significant role in shaping the overall coherence and quality of the output compared to later tokens (c.f. Appendix~\ref{sec:dynamic_decode}). Thus, we propose \emph{Progressive Low-rank Decoding}, a dynamic compression strategy that gradually reduces the model's overall activated ranks during decoding. As shown in Figure~\ref{fig:progressive}, our method progressively decreases the rank as more tokens are generated, increasing the overall compression rate while preserving strong performance in generation phase.

To adapt the rank allocation during decoding, we design a scheduler that determines the overall rank budget \(R_{\text{budget}}\) to be used for each token. Our scheduler leverages a calibration dataset to identify the optimal schedule based on different target compression levels. Let \(R_{\text{budget}}(t)\) denote the rank budget for token \(t\) as determined by the scheduler. Note that \(R_{\text{budget}}(t)\) is non-increasing, meaning that while consecutive tokens may share the same budget, the budget for token \(t+1\) will never exceed that for token \(t\).

Substituting \(R_{\text{budget}}(t)\) for \(R_{\text{budget}}\) in Equation~\ref{eq:rank_allocation} yields the token-specific rank configuration:
\begin{equation}
r_{l,p}(t) \;=\; \mathrm{round}\Bigl(\frac{\alpha_{l,p}}{S} \times R_{\text{budget}}(t)\Bigr).
\end{equation}
This yields the configuration \(\{r_{l,p}(t) \mid l \in \mathcal{L},\, p \in P_l\}\) for the current token.

This scheduler-based approach dynamically adjusts the rank budget during decoding: early tokens benefit from a larger parameter set, while later tokens are generated with a reduced rank configuration. For supplementary details on our method, please refer to Appendix~\ref{sec:pgl}.

\begin{table*}[]
\resizebox{\textwidth}{!}{%
\begin{tabular}{@{}c|c|cccc|cccc@{}}
\toprule
\multirow{3}{*}{Comp. Rate} &
  \multirow{3}{*}{Method} &
  \multicolumn{4}{c|}{Llama-3-8B-Instruct} &
  \multicolumn{4}{c}{Llama-2-7B-Chat} \\
 &      & \multicolumn{2}{c}{DialogSum} & \multicolumn{2}{c|}{CNN/DM} & \multicolumn{2}{c}{DialogSum} & \multicolumn{2}{c}{CNN/DM} \\
 &
   &
  ROUGE-L ↑ &
  BERTScore ↑ &
  ROUGE-L ↑ &
  BERTScore ↑ &
  ROUGE-L ↑ &
  BERTScore ↑ &
  ROUGE-L ↑ &
  BERTScore ↑ \\ \midrule
- &
  Baseline &
  24.72 &
  86.79 &
  24.34 &
  86.51 &
  24.56 &
  87.75 &
  24.82 &
  87.23 \\ \midrule
\multirow{3}{*}{20\%} &
  ASVD &
  0.10 &
  80.07 &
  0.54 &
  77.09 &
  15.44 &
  80.45 &
  7.94 &
  78.75 \\
 &
  SVD-LLM &
  0.24 &
  78.12 &
  6.29 &
  76.46 &
  13.62 &
  83.07 &
  19.71 &
  \textbf{84.86} \\
 &
  FLRC &
  \textbf{17.35} &
  \textbf{86.00} &
  \textbf{17.72} &
  \textbf{84.18} &
  \textbf{17.22} &
  \textbf{85.29} &
  \textbf{19.84} &
  84.83 \\ \midrule
\multirow{3}{*}{30\%} &
  ASVD &
  0.53 &
  72.45 &
  0.07 &
  71.81 &
  6.47 &
  80.34 &
  3.44 &
  75.66 \\
 &
  SVD-LLM &
  0.41 &
  72.06 &
  3.98 &
  74.28 &
  2.34 &
  75.62 &
  15.56 &
  82.20 \\
 & FLRC & \textbf{8.09}     & \textbf{81.92}     & \textbf{10.83}    & \textbf{79.92}   & \textbf{14.91}     & \textbf{83.62}    & \textbf{17.28}   & \textbf{83.91}   \\ \bottomrule
\end{tabular}%
}
\caption{Generative performance comparison (ROUGE-L and BertScore are expressed as percentages).}
\label{tab:rouge}
\end{table*}
\begin{table*}[htb]
\resizebox{\textwidth}{!}{%
\begin{tabular}{@{}c|c|c|c|cccccccc@{}}
\toprule
\multirow{2}{*}{Model} &
  \multirow{2}{*}{Comp. Rate} &
  \multirow{2}{*}{Method} &
  Perplexity ↓ &
  \multicolumn{8}{c}{Zero-shot Task Accuracy (\%) ↑} \\
                            &                       &          & Wiki2 & ARC-e          & ARC-c          & Hella & OBQA  & Wino  & MathQA         & PIQA  & Avg.  \\ \midrule
\multirow{7}{*}{Llama-3-8B} & -                   & Baseline & 6.14      & 80.13          & 50.51          & 60.17 & 34.80 & 72.61 & 40.50          & 79.71 & 59.78 \\ \cmidrule(l){2-12} 
                            & \multirow{3}{*}{20\%} & ASVD     & 3206.80   & 30.81          & 19.54          & 27.06 & 13.80 & 52.41 & 21.04          & 56.37 & 31.58 \\
                            &                       & SVD-LLM  & 14.72     & \textbf{55.64} & 27.30          & 37.22 & 21.60 & 60.54 & 24.39          & 64.69 & 41.63 \\
 &
   &
  FLRC &
  \textbf{12.53} &
  54.42 &
  \textbf{28.58} &
  \textbf{38.95} &
  \textbf{23.80} &
  \textbf{68.27} &
  \textbf{25.03} &
  \textbf{66.54} &
  \textbf{43.66} \\ \cmidrule(l){2-12} 
                            & \multirow{3}{*}{30\%} & ASVD     & 28566.03  & 25.58          & \textbf{22.78} & 25.84 & 12.40 & 51.22 & 18.26          & 52.29 & 29.77 \\
                            &                       & SVD-LLM  & 33.13     & \textbf{40.07} & 20.99          & 30.30 & 16.80 & 55.33 & \textbf{22.75} & 57.94 & 34.88 \\
 &
   &
  FLRC &
  \textbf{25.46} &
  38.34 &
  20.39 &
  \textbf{30.84} &
  \textbf{19.00} &
  \textbf{59.51} &
  21.68 &
  \textbf{60.55} &
  \textbf{35.76} \\ \bottomrule
\end{tabular}%
}
\caption{Perplexity and zero-shot accuracy of low-rank compression methods.}
\label{tab:ppl}
\end{table*}

\section{Experiments}
\subsection{Experiments Setup}
For the decoding stage evaluation, we conduct experiments on two summarization benchmarks: DialogSum \citep{chen2021dialogsum} and CNN/DM \citep{hermann2015teaching}. In addition, to assess performance during the prefilling stage, we measure the perplexity on the Wikitext2 \citep{merity2016pointer} dataset and evaluate zero-shot accuracy across seven common tasks provided in the LM-Evaluation-Harness \citep{gao2021framework}. For experimental details, please refer to the Appendix~\ref{sec:details}.

\subsection{Evaluation on Generation Tasks}
As shown in Table~\ref{tab:rouge}, our experiments on Llama-3-8B-Instruct \citep{dubey2024llama} reveal that previous low-rank compression methods struggle with generation tasks. In contrast, our approach, which incorporates progressive low-rank decoding, consistently maintains strong performance across various compression ratios. Here, the compression rate represents the overall percentage of parameter usage saved during the entire generation stage. Notably, under a 20\% compression rate\footnote{We define the compression rate as the average percentage reduction in model parameters per token, computed over both the prefilling and decoding stages.}, evaluations on the DialogSum benchmark indicate that while competing methods yield ROUGE-L scores of less than 1\%, our method achieves an impressive 17.35\%.

Although earlier low-rank compression techniques have shown relatively better performance on Llama-2-7B-Chat, our method still delivers significantly higher ROUGE-L and BertScore metrics at high compression rates across both benchmarks. An ablation study of our proposed approach is presented in Appendix~\ref{sec:ablation}. We also evaluated our method on different model sizes to demonstrate its generalization; see Appendix~\ref{sec:diff_size} for details.

\subsection{Evaluation on Understanding Tasks}
In addition to generation tasks, we evaluate our approach on perplexity and zero-shot accuracy using Llama-3-8B, as shown in Table~\ref{tab:ppl}. On the Wikitext2 dataset, our method achieves significantly lower perplexity compared to other low-rank compression techniques. Moreover, the average zero-shot accuracy across various compression ratios consistently outperforms that of previous methods. These results indicate that our proposed layer-wise rank allocation effectively mitigates the performance loss typically associated with model compression, ensuring robust language understanding even under aggressive parameter reduction.

\subsection{Rank Allocation Search Time}
\label{exp:search_time}
We compare our proposed rank allocation search with the ASVD approach. The ASVD method, being perplexity-based, requires substantially more time for the search process compared to our approach. On an A100 GPU, the ASVD method takes approximately 147 minutes to complete the search, whereas our method requires only 3 minutes, representing a 49-fold improvement in speed. This significant reduction in search time demonstrates that our approach can quickly and efficiently determine an optimal rank configuration for the model, thereby facilitating faster deployment. For additional performance comparison experiments, please refer to Appendix~\ref{sec:rank_ablation}.

\section{Conclusion}
In this study, we propose the \emph{Fine-grained Low-Rank Compressor (FLRC)} to rapidly determine the optimal compression ratio for each layer, thereby mitigating the performance degradation that arises from applying a uniform compression rate across all layers. Additionally, we introduce progressive low-rank decoding to address the poor performance of existing low-rank compression methods during the generation phase. Experimental results demonstrate that, under the same parameter utilization, our approach outperforms other methods on both generation and understanding tasks, indicating a significant performance improvement in low-rank compression.
\section*{Limitation}
In this study, we rely on a calibration dataset to perform layer-wise rank allocation and design the scheduler for \emph{FLRC}. However, the model's performance on different benchmarks may vary depending on the choice of calibration dataset, which can lead to discrepancies. To ensure fairness, we use the same calibration dataset for all methods in our experiments.

Additionally, our experimental results show that dynamically specifying the number of model parameters used per token can greatly enhance LLM inference efficiency. Nevertheless, optimizing the scheduler for dynamic rank allocation remains a crucial challenge, as it may introduce additional overhead. Consequently, our future work will focus on engineering optimizations and kernel design, specifically reducing the overhead associated with dynamic rank allocation, to further improve the overall efficiency of our approach.
\section*{Acknowledgment}
We would like to express our gratitude to all organizations that provided the computational resources necessary to complete the experiments in this study. Additionally, we acknowledge the use of ChatGPT for assisting with paraphrasing and polishing, and not for any other illegal purposes.

\bibliography{custom}

\appendix
\section{Importance Score of Different Layers}
\label{sec:importance}

\begin{figure}[htb]
  \includegraphics[width=\columnwidth]{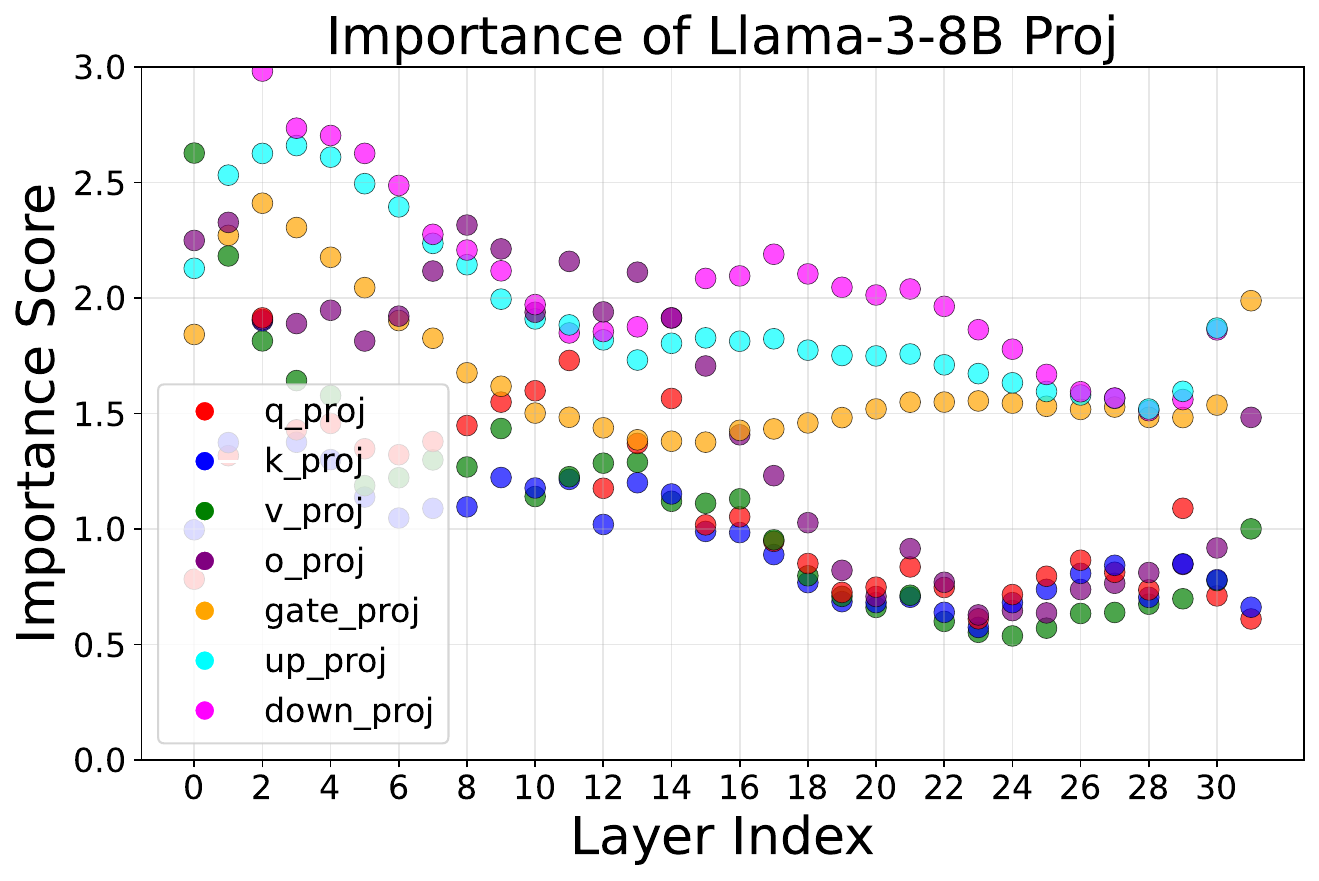}
  \caption{The importance score of various projections in Llama-3-8B across different layer indices. Each point represents a projection's score; higher scores (e.g., "down\_proj") indicate that less compression should be applied, while lower scores allow for more aggressive compression.}
  \label{fig:importance}
\end{figure}

Different layers within a model often exhibit varying degrees of “compressibility”, implying that uniform compression ratio can lead to suboptimal results. We can calculate the importance score of each component in the model based on our proposed method. As shown in Figure~\ref{fig:importance}, the importance score of the projection in each layer varies significantly. Identifying which layers can tolerate more aggressive compression and which layers require a more careful approach is crucial to maximizing efficiency while minimizing performance degradation.

\section{Sensitivity Metrics for Each Projection}
\label{sec:metrics}

We use a small calibration dataset and perform back propagation to compute the gradient for each projection. We observed that parameters with larger gradients tend to be more sensitive, and that larger weight values typically indicate higher importance. Thus, we multiply the weight and its corresponding gradient and then square the product to derive an importance value.

In addition to Equation~\ref{eq:alpha}, we evaluated two alternative metrics. First, we considered only the weight magnitudes:
\begin{equation}
\label{eq:weight}
\alpha_{l,p} \;=\; \sum_{i}\Bigl( \mathbf{W}_{l,p}[i]\Bigr)^2,
\end{equation}
and second, we considered only the gradient values:
\begin{equation}
\label{eq:grad}
\alpha_{l,p} \;=\; \sum_{i}\Bigl(\mathbf{G}_{l,p}[i]\Bigr)^2.
\end{equation}
Using each metric, we computed the importance of every projection and performed rank allocation accordingly. Table~\ref{tab:metrics_performance} presents generative performance comparison on Llama-3-8B-Instruct. It is clear that the metric combining both gradient and weight magnitudes is the most accurate. Consequently, we adopt Equation~\ref{eq:alpha} as our chosen method for estimating the importance of projection.

\begin{table}[]
\centering
\resizebox{0.9\columnwidth}{!}{%
\begin{tabular}{@{}c|c|c@{}}
\toprule
Comp. Rate            & Method   & DialogSum ROUGE-L ↑ \\ \midrule
-                     & Baseline & 24.72               \\ \midrule
\multirow{3}{*}{10\%} & Eq.~\ref{eq:weight}    & 0.44                \\
                      & Eq.~\ref{eq:grad}    & 15.74               \\
                      & Eq.~\ref{eq:alpha}     & \textbf{20.23}      \\ \midrule
\multirow{3}{*}{20\%} & Eq.~\ref{eq:weight}    & 0.07                \\
                      & Eq.~\ref{eq:grad}    & 2.16                \\
                      & Eq.~\ref{eq:alpha}    & \textbf{17.35}      \\ \bottomrule
\end{tabular}%
}
\caption{Generative performance comparison of different sensitivity metrics on Llama-3-8B-Instruct (ROUGE-L is expressed as percentages).}
\label{tab:metrics_performance}
\end{table}

\section{Supplementary Details on Progressive Low-Rank Decoding}
\label{sec:pgl}

Increasing the compression rate gradually during the generation phase is highly compatible with low-rank compression. After decomposing each projection’s parameter matrix into two smaller matrices using singular value decomposition, the channels in these matrices are automatically ordered by importance. In other words, rows or columns at lower indices contain the most critical information, while those at higher indices can be safely truncated. This allows us to dynamically decide, at each token generation step, how many of the top \textit{k} rows or columns to retain, where a smaller \textit{k} corresponds to a higher compression rate. This inherent property makes our approach ideally suited for dynamic rank allocation, leading to an efficient implementation of progressive low-rank decoding, as we only need to decrement \textit{k} during token generation.

Although dynamic rank adjustment introduces some overhead, when the savings in computation and data transfer are substantial, the overhead of dynamically changing the rank becomes negligible. Moreover, users can also evaluate what level of performance degradation is acceptable in exchange for the corresponding acceleration, as this will vary depending on the specific use case.

In our work, the term “schedule” refers to the points during the generation process at which the LLM switches to a higher compression rate. This means we can generate many schedule candidates, each corresponding to an overall compression rate (i.e., the average compression rate used for every token), which we denote as the overall rank budget ($R_{\text{budget}}$). We then use a calibration dataset to evaluate the performance of each schedule (using metrics like BERTScore), ultimately selecting the schedule that best meets our desired overall rank budget $R_{\text{budget}}$ while achieving optimal performance for running our FLRC.
\section{Progressive Low-rank Decoding Forms}
\label{sec:dynamic_decode}

In this study, we propose dynamically adjusting the number of ranks used for each generated token, and we compare three approaches for doing so. The first approach, \emph{Static Rank Decoding}, applies a fixed rank for every token. The second, \emph{Increased Rank Decoding}, uses fewer ranks for early tokens and more for later ones. The third, \emph{Decreased Rank Decoding}, assigns more ranks to early tokens and fewer to later tokens. In Table~\ref{tab:dynamic_rank}, we compare these methods on the DialogSum summarization task, ensuring that each approach uses the same average number of parameters. Our results demonstrate that \emph{Decreased Rank Decoding} achieves superior performance, which is why we adopt it as our method for \emph{Progressive Low-Rank Decoding}.


\begin{table}[]
\resizebox{\columnwidth}{!}{%
\begin{tabular}{@{}c|c@{}}
\toprule
Method                  & DialogSum ROUGE-L ↑ \\ \midrule
Static rank decoding    & 14.71               \\
Increased rank decoding & 8.59                \\
Decreased rank decoding & \textbf{19.87}      \\ \bottomrule
\end{tabular}%
}
\caption{Comparison of different dynamic rank decoding methods on Llama-3-8B-Instruct (ROUGE-L is expressed as percentages).}
\label{tab:dynamic_rank}
\end{table}
\section{Ablation Study}
\label{sec:ablation}

Our proposed Fine-grained Low-Rank Compressor consists of two key components: Fisher-based Layer-wise Rank Allocation (FLRA) and Progressive Low-Rank Decoding (PLRD). To quantify the impact of each component, we conducted an ablation study on Llama-3-8B-Instruct with a 20\% compression rate, measuring generative performance. As shown in Table~\ref{tab:ablation}, SVD-LLM alone delivers poor results. In contrast, applying either our FLRA or PLRD individually yields substantial gains in generation quality. These findings demonstrate that both components of our method effectively enhance the performance of low-rank compressed models.

\newcommand{\cmark}{\ding{51}}  
\newcommand{\xmark}{\ding{55}}  

\begin{table}[]
\resizebox{\columnwidth}{!}{%
\begin{tabular}{@{}ccc|c@{}}
\toprule
\multicolumn{3}{c|}{Ablation Settings}                                            & \multirow{2}{*}{DialogSum ROUGE-L ↑} \\
SVD-LLM & FLRA & PLRD &                            \\ \midrule
\cmark   & \xmark                                   & \xmark                         & 0.24                       \\
\cmark   & \cmark                                   & \xmark                         & 13.28                      \\
\cmark   & \cmark                                   & \cmark                         & \textbf{17.35}                      \\ \bottomrule
\end{tabular}%
}
\caption{Ablation study on generative performance (ROUGE-L is expressed as percentages). “FLRA” denotes Fisher-based Layer-wise Rank Allocation, and “PLRD” denotes Progressive Low-Rank Decoding.}
\label{tab:ablation}
\end{table}

\section{Experimental Details}
\label{sec:details}

Our used datasets and base models were sourced from the HuggingFace \citep{lhoest2021datasets} and Transformers \citep{wolf2020transformers} libraries, and all usage complied with the respective terms and conditions. For evaluating zero-shot accuracy, we employed seven common tasks: ARC-Easy, ARC-Challenge \citep{Clark2018ThinkYH}, HellaSwag \citep{zellers2019hellaswag}, OpenBookQA \citep{OpenBookQA2018}, WinoGrande \citep{sakaguchi2019winogrande}, MathQA \citep{amini2019mathqa} and PIQA \citep{Bisk2020}. For summarization tasks, we used ROUGE-L \citep{lin2004rouge} and BertScore \citep{zhang2019bertscore} as evaluation metrics.

For the \emph{FLRC} layer-wise rank allocation, we sampled 256 sequences (each with a length of 2048) from the Wikitext2 training set as our calibration dataset, while the scheduler's calibration dataset was drawn from 500 samples from the DialogSum training set. For perplexity evaluation, the input sequence length was set to 2048. The compression rate is computed by first establishing a baseline based on the number of parameters in the \texttt{q\_proj}, \texttt{k\_proj}, \texttt{v\_proj}, \texttt{o\_proj}, \texttt{gate\_proj}, \texttt{up\_proj}, and \texttt{down\_proj} matrices of the LLaMA model, and then determining the percentage of parameters omitted during each inference.

Our experimental pipeline follows the SVD-LLM procedure. First, the model weights are decomposed using SVD-LLM's truncation-aware data whitening method, after which we apply our proposed layer-wise rank allocation and progressive low-rank decoding modules. Notably, since our compression strategy is orthogonal to PEFT fine-tuning, we deliberately omit the weight updating steps typically included in the SVD-LLM framework. This design choice was made to ensure a fair comparison with SVD-LLM.
\section{Rank Allocation Method Comparison}
\label{sec:rank_ablation}

In order to compare our rank allocation method with ASVD's, we first whiten the model weights and then apply different rank allocation strategies. We evaluate the resulting models on Wikitext2 by measuring perplexity. As shown in Table~\ref{tab:rank_ablation}, under the same compression rate, our method achieves lower perplexity, demonstrating that our approach not only speeds up the search process but also finds a more optimal rank allocation for the compressed model.

\begin{table}[h]
\resizebox{\columnwidth}{!}{%
\begin{tabular}{@{}c|c|c@{}}
\toprule
Comp. Rate            & Rank Allocation Method & Wiki2 Perplexity ↓ \\ \midrule
\multirow{2}{*}{20\%} & ASVD                   & 22.69              \\
                      & FLRC                   & \textbf{12.53}     \\ \midrule
\multirow{2}{*}{30\%} & ASVD                   & 128.96             \\
                      & FLRC                   & \textbf{25.46}     \\ \bottomrule
\end{tabular}%
}
\caption{Rank allocation method comparison on Llama-3-8B.}
\label{tab:rank_ablation}
\end{table}

The rank allocation method we employ is both fast (as detailed in Section~\ref{exp:search_time}) and yields superior results. Unlike techniques that rely on iterative updates (such as Bayesian Optimization \citep{ji2024feature}) or memory-intensive and slow Hessian-based methods \citep{shao2024one}, our approach avoids these drawbacks.

Previous works \citep{lin2024modegpt, ji2024feature, shao2024one} have proposed estimating the importance of various model components; however, these approaches are often inefficient or inaccurate and unsuitable for our method. MoDeGPT  \citep{lin2024modegpt} evaluates the importance of different blocks (or layers) using block influence, which requires the input and output dimensions to be the same. In contrast, our rank allocation method is more fine-grained and can evaluate the importance of each projection within every block, making it better suited for our progressive low-rank decoding. Bolaco \citep{ji2024feature} uses Bayesian optimization for rank allocation, which requires multiple iterations to converge. Our approach, on the other hand, only needs a single iteration, making it significantly more efficient. PrunerGPT \citep{shao2024one} uses a Hessian-based approach to identify the importance of each component, which consumes substantial memory and computation time. As a result, these methods are less efficient than our proposed method.

We integrated the allocation methods from prior works with our progressive low-rank decoding and conducted a generative performance comparison. As shown in Table~\ref{tab:diff_allocation}, our fisher-based rank allocation outperforms the other methods and remains highly efficient.

\begin{table}[]
\resizebox{\columnwidth}{!}{%
\begin{tabular}{@{}c|c|c@{}}
\toprule
Comp. Rate            & Rank Allocation Method & Dialogsum ROUGE-L ↑ \\ \midrule
-                     & Baseline               & 24.56               \\ \midrule
\multirow{3}{*}{20\%} & MoDeGPT                & 3.91                \\
                      & PrunerGPT              & 16.28               \\
                      & \textbf{FLRC}          & \textbf{17.22}      \\ \midrule
\multirow{3}{*}{30\%} & MoDeGPT                & 2.43                \\
                      & PrunerGPT              & 10.81               \\
                      & \textbf{FLRC}          & \textbf{14.91}      \\ \bottomrule
\end{tabular}%
}
\caption{Generative performance comparison of different allocation methods on Llama-2-7B-Chat (ROUGE-L is expressed as percentages).}
\label{tab:diff_allocation}
\end{table}

\section{Evaluation on Models of Different Sizes}
\label{sec:diff_size}

We evaluated our method on 3B and 13B models to demonstrate its generalization capability. Table~\ref{tab:diff_size} clearly shows that FLRC continues to outperform SVD-LLM by a significant margin. The results demonstrate that, although SVD-LLM experiences a significant performance drop, FLRC substantially mitigates the performance degradation at the same compression rate.

\begin{table}[h]
\resizebox{\columnwidth}{!}{%
\begin{tabular}{@{}c|c|cc@{}}
\toprule
\multirow{2}{*}{Method} & \multirow{2}{*}{Comp. Rate} & \multicolumn{2}{c}{DialogSum ROUGE-L ↑} \\
         &                       & Llama3.2-3B                         & Llama-2-13B    \\ \midrule
Baseline & -                     & \multicolumn{1}{c|}{12.84}          & 17.23          \\ \midrule
SVD-LLM  & \multirow{2}{*}{10\%} & \multicolumn{1}{c|}{7.09}           & 16.94          \\
FLRC     &                       & \multicolumn{1}{c|}{\textbf{13.98}} & \textbf{17.99} \\ \midrule
SVD-LLM  & \multirow{2}{*}{20\%} & \multicolumn{1}{c|}{3.55}           & 0.18           \\
FLRC     &                       & \multicolumn{1}{c|}{\textbf{9.94}}  & \textbf{17.43} \\ \bottomrule
\end{tabular}%
}
\caption{Generative performance comparison on 3B and 13B models (ROUGE-L is expressed as percentages).}
\label{tab:diff_size}
\end{table}

We also conducted a zero-shot evaluation on the Llama-2-13B model. As shown in Table~\ref{tab:13b_zeroshot}, our method consistently outperforms prior approaches across diverse tasks at the same compression rate, highlighting the superiority of FLRC efficacy in preserving models performance.

\begin{table*}[h]
\centering
\resizebox{0.8\linewidth}{!}{%
\begin{tabular}{@{}c|c|cccccccc@{}}
\toprule
\multirow{2}{*}{Comp. Rate} & \multirow{2}{*}{Method} & \multicolumn{8}{c}{Zero-shot Task Accuracy (\%) ↑} \\
 &  & ARC-e & ARC-c & Hella & OBQA & Wino & MathQA & PIQA & Avg. \\ \midrule
- & Baseline & 79.42 & 48.29 & 60.05 & 35.20 & 72.30 & 32.13 & 79.05 & 58.06 \\ \midrule
\multirow{2}{*}{20\%} & SVD-LLM & 67.89 & 32.76 & 44.28 & 29.00 & 67.56 & 25.59 & 71.11 & 48.31 \\
 & FLRC & \textbf{70.33} & \textbf{38.05} & \textbf{47.49} & \textbf{31.20} & \textbf{69.53} & \textbf{27.91} & \textbf{72.91} & \textbf{51.06} \\ \midrule
\multirow{2}{*}{30\%} & SVD-LLM & 58.71 & 25.94 & 37.80 & \textbf{26.60} & 64.56 & 24.49 & 66.54 & 43.52 \\
 & FLRC & \textbf{63.64} & \textbf{30.12} & \textbf{41.52} & \textbf{26.60} & \textbf{66.14} & \textbf{24.72} & \textbf{68.39} & \textbf{45.88} \\ \bottomrule
\end{tabular}%
}
\caption{Zero-shot comparison results on Llama2-13B.}
\label{tab:13b_zeroshot}
\end{table*}

We further evaluated our approach on the Llama-30B model (i.e., models exceeding 20B parameters), as presented in Table~\ref{tab:30b_exp}. On this larger scale, our method continues to outperform prior techniques at identical compression rates. Moreover, we observe that our technique achieves even greater compression efficiency on larger models, yielding a smaller accuracy drop.

\begin{table}[h]
\centering
\resizebox{0.8\columnwidth}{!}{%
\begin{tabular}{@{}c|c|c|c@{}}
\toprule
Comp. Rate & Method & \begin{tabular}[c]{@{}c@{}}Wiki2 \\ Perplexity ↓\end{tabular} & \begin{tabular}[c]{@{}c@{}}Dialogsum \\ Rouge-L ↑\end{tabular} \\ \midrule
- & Baseline & 4.10 & 17.25 \\ \midrule
\multirow{2}{*}{20\%} & SVD-LLM & 5.55 & 16.77 \\
 & FLRC & \textbf{5.21} & \textbf{18.95} \\ \midrule
\multirow{2}{*}{30\%} & SVD-LLM & 6.27 & 16.31 \\
 & FLRC & \textbf{5.75} & \textbf{18.98} \\ \midrule
\multirow{2}{*}{40\%} & SVD-LLM & 7.58 & 0.00 \\
 & FLRC & \textbf{6.62} & \textbf{18.19} \\ \bottomrule
\end{tabular}%
}
\caption{Performance comparison on 30B model.}
\label{tab:30b_exp}
\end{table}
\section{Speedup of End-to-end Decoding}
\label{sec:speedup}

We conducted practical speedup experiments on our method. Table~\ref{tab:speedup} is our current acceleration result using the Llama-3-8B-Instruct model with a batch size of 512, a sequence length of 32, and 128 tokens generated. These results still show a tangible speedup. Typically, benchmarks for such work increase the batch size to make the model compute-bound and achieve higher throughput.

\begin{table}[h]
\resizebox{\columnwidth}{!}{%
\begin{tabular}{@{}c|c|c|c@{}}
\toprule
Method                & Comp. Rate & \begin{tabular}[c]{@{}c@{}}Throughput (tokens/sec)\end{tabular} & Speedup \\ \midrule
Baseline              & -          & 3646.62                 & 1x      \\ \midrule
\multirow{3}{*}{FLRC} & 20\%       & 3856.99                 & 1.06x   \\
                      & 30\%       & 4051.53                 & 1.11x   \\
                      & 40\%       & 5290.33                 & 1.45x   \\ \bottomrule
\end{tabular}%
}
\caption{Speedup of FLRC on Llama-3-8B-Instruct.}
\label{tab:speedup}
\end{table}

However, we believe that our proposed progressive low-rank decoding is particularly effective for alleviating memory-bound issues as well as situations characterized by low throughput. To further validate this, we conducted an additional experiment under offloading conditions. In this setup, using the same Llama-3-8B-Instruct model, our GPU is limited to approximately 8GB of VRAM; hence, the remaining parameter matrices are offloaded to host DRAM and transferred to GPU VRAM when needed for computation. The experimental settings in this case are: a batch size of 1, sequence length of 32, and generating 128 tokens. Table~\ref{tab:offload} is our experimental result for offloading. Our approach alleviates the memory transfer requirements, thereby accelerating the overall process. Our results clearly demonstrate that our method yields even more significant acceleration when the system is memory-bound. Additionally, in data transfers, larger data tend to experience increased fragmentation compared to smaller ones. This fragmentation means that the data is divided into more segments or fragments, and each fragment often incurs its own processing overhead. Therefore, in strongly memory-bound situations, FLRC may deliver even better acceleration than theoretically predicted.

\begin{table}[h]
\centering
\resizebox{\columnwidth}{!}{%
\begin{tabular}{@{}c|c|c|c@{}}
\toprule
Method & Comp. Rate & \begin{tabular}[c]{@{}c@{}}Throughput (tokens/sec)\end{tabular} & Speedup \\ \midrule
Baseline              & -    & 1.20 & 1x    \\ \midrule
\multirow{3}{*}{FLRC} & 20\% & 1.40 & 1.17x \\
                      & 30\% & 1.83 & 1.53x \\
                      & 40\% & 2.54 & 2.12x \\ \bottomrule
\end{tabular}%
}
\caption{Offloading speedup of FLRC on Llama-3-8B-Instruct.}
\label{tab:offload}
\end{table}

As models grow larger and context windows increase, GPU VRAM demand will rise, making offloading scenarios increasingly common for single user and edge device. Under these conditions, the model's inherent throughput can become very low. Therefore, FLRC is particularly beneficial in environments that require model offloading.
\section{FLRC on Low-precision Model}
\label{sec:lowbit}
We conducted our experiments primarily in FP16 precision. As shown in Table~\ref{tab:lowbit}, our method remains equally effective at lower precisions. We evaluated generation task on both Llama-3-8B-Instruct and Llama-2-7B-Chat models using our approach. The results demonstrate that there is no drop in accuracy across various compression rates, even when using lower-precision models. This confirms that our parameter-reduction technique and low-precision quantization work synergistically.

\begin{table}[]
\centering
\resizebox{\columnwidth}{!}{%
\begin{tabular}{@{}c|c|cc@{}}
\toprule
\multicolumn{1}{l|}{\multirow{2}{*}{Comp. Rate}} & \multicolumn{1}{l|}{\multirow{2}{*}{Precision}} & \multicolumn{2}{c}{Dialogsum Rouge-L ↑} \\
\multicolumn{1}{l|}{} & \multicolumn{1}{l|}{} & Llama-3-8B-Instruct & Llama-2-7B-Chat \\ \midrule
- & FP16 & 24.72 & 24.56 \\ \midrule
\multirow{2}{*}{20\%} & FP16 & 17.35 & 17.22 \\
 & INT8 & 17.48 & 17.47 \\ \midrule
\multirow{2}{*}{30\%} & FP16 & 8.09 & 14.91 \\
 & INT8 & 7.81 & 15.19 \\ \bottomrule
\end{tabular}%
}
\caption{Generative performance comparison on low-precision models.}
\label{tab:lowbit}
\end{table}

\section{Sensitivity Analysis to Calibration Datasets}
\label{sec:calib}

Most existing SVD-based methods rely on calibration datasets. Table~\ref{tab:calib} shows experimental results obtained by calibrating on different datasets for compressing Llama-3.2-3B. Notably, when calibrated on Wikitext2, the model exhibits improved perplexity on Wikitext2 but performs worse on C4; conversely, calibration on C4 yields better results on C4 but poorer performance on Wikitext2. This behavior is expected, as models tend to perform better on data that closely resembles the calibration set. Importantly, our results indicate that FLRC consistently achieves lower perplexity than SVD-LLM across different calibration datasets. Therefore, as long as all compared methods are calibrated using the same dataset, the experiments remain fair. In all our experiments, FLRC and the previous methods (ASVD, SVD-LLM) have been calibrated on the identical dataset.

\begin{table*}[!t]
\centering
\resizebox{0.7\linewidth}{!}{%
\begin{tabular}{@{}c|c|cc|cc@{}}
\toprule
\multirow{2}{*}{Method} & \multirow{2}{*}{Comp. Rate} & \multicolumn{2}{c|}{Calibration on Wikitext2} & \multicolumn{2}{c}{Calibration on C4} \\
         &                       & Wiki2 ↓        & C4 ↓           & Wiki2 ↓        & C4 ↓           \\ \midrule
Baseline & -                     & 7.81           & 11.33          & 7.81           & 11.33          \\ \midrule
SVD-LLM  & \multirow{2}{*}{10\%} & 14.72          & 48.03          & 38.01          & 29.63          \\
FLRC     &                       & \textbf{11.39} & \textbf{25.79} & \textbf{18.99} & \textbf{18.55} \\ \midrule
SVD-LLM  & \multirow{2}{*}{20\%} & 26.95          & 120.92         & 117.74         & 53.78          \\
FLRC     &                       & \textbf{19.12} & \textbf{58.92} & \textbf{42.92} & \textbf{27.41} \\ \bottomrule
\end{tabular}%
}
\caption{Perplexity on different calibration datasets on Llama-3.2-3B.}
\label{tab:calib}
\end{table*}

\end{document}